\documentclass[conference]{IEEEtran}
\IEEEoverridecommandlockouts

\usepackage{amsmath,amssymb,amsfonts}
\usepackage{graphicx}
\usepackage{booktabs,makecell,multirow,tabularx,array}
\usepackage{adjustbox}
\usepackage{xcolor}
\usepackage{pifont}       
\usepackage{placeins}
\usepackage{float}
\usepackage{cite}

\usepackage{algorithm}
\usepackage[noend]{algpseudocode}
\algrenewcommand\algorithmicrequire{\textbf{Input:}}
\algrenewcommand\algorithmicensure{\textbf{Output:}}
\algnewcommand{\LineComment}[1]{\State \(\triangleright\) #1}

\newcommand{\cmark}{\ding{51}}
\newcommand{\xmark}{\ding{55}}

\usepackage{textcomp}
\usepackage{xcolor}
\usepackage{graphicx}

\usepackage{float}

\usepackage{array,tabularx,booktabs}
\usepackage{placeins}      
\usepackage{caption}

\usepackage{hyperref}

\begin{document}

\title{Sequence-Preserving Dual-FoV Defense for Traffic Sign and Light Recognition in Autonomous Vehicles
}

\author{\IEEEauthorblockN{Abhishek Joshi}
\IEEEauthorblockA
{\textit{Department of Computer Science} \\
\textit{Texas A\&M University-Corpus Christi}\\
Corpus Christi, USA \\
ajoshi5@islander.tamucc.edu}
\and
\IEEEauthorblockN{ Jahnavi Krishna Koda}
\IEEEauthorblockA
{\textit{Coastal and Marine System Science} \\
\textit{Texas A\&M University-Corpus Christi}\\
Corpus Christi, USA \\
jkoda@islander.tamucc.edu}
\and
\IEEEauthorblockN{ Abhishek Phadke}
\IEEEauthorblockA
{\textit{School of Engineering \& Computing} \\
\textit{Christopher Newport University}\\
Virginia, USA \\
abhishek.phadke@cnu.edu}
\and

}
\maketitle

\begin{abstract}
Traffic light and sign recognition are key for Autonomous Vehicles (AVs) because perception mistakes directly influence navigation  and safety. In addition to digital adversarial attacks, models are vulnerable to existing perturbations (glare, rain, dirt, or graffiti), which could lead to dangerous misclassifications. The current work lacks consideration of temporal continuity, multistatic field-of-view (FoV) sensing, and robustness to both digital and natural degradation. This study proposes a dual FoV, sequence-preserving robustness framework for traffic lights and signs in the USA based on a multi-source dataset built on aiMotive, Udacity, Waymo, and self-recorded videos from the region of Texas. Mid and long-term sequences of RGB images are temporally aligned for four operational design domains (ODDs): highway, night, rainy, and urban. Over a series of experiments on a real-life application of anomaly detection, this study outlines  a unified three-layer defense stack framework that incorporates feature squeezing, defensive distillation, and entropy-based anomaly detection, as well as sequence-wise temporal voting for further enhancement. The evaluation measures included accuracy, attack success rate (ASR), risk-weighted misclassification severity, and confidence stability. Physical transferability was confirmed using probes for recapture. The results showed that the Unified Defense Stack achieved 79.8\% mAP and reduced the ASR to 18.2\% (from baseline 37.4\%), which is superior to YOLOv8, YOLOv9, and BEVFormer, while reducing the high-risk misclassification to 32\%.
\end{abstract}


\section{Introduction}
\label{sec:intro}
The perception of AV is camera-based, and it allows machines to understand and see two essential concepts: traffic lights and traffic signs, after which further planning and control decisions are made.  

However, perception models are uncertain when subjected to pertinences arising naturally or due to physical conditions during deployment. These do not require an attacker to gain digital access to the AV, but could be caused by real-life conditions such as laser glare, dirt, raindrops landing on the lens, a roll of sunshine, graffiti, or stickers affixed to signs. Although safety-critical errors can be caused by even minor distortions, such as a `` Stop' sign misread as a sign with a speed limit of 60, or even a green light interpreted as a red one, these errors occur in real-world deployments and software testing (notably, fuzzing) \cite{eykholt2018robust,zhang2021evaluating}. 
The original work with adversarial examples in computer vision was demonstrated by \cite{goodfellow2015explaining} and explained later by \cite{eykholt2018robust}, followed by applications to traffic signs in the real world and to the full interactions of AV perception \cite{cao2021simple,xiong2021multisource}. Partial protection was provided by the use of defensive distillation ( defense) \cite{papernot2016distillation} and certified robustness \cite{cohen2019certified} but generally favored accuracy guarantees over guarantees and could not withstand hybrid or unexpected perturbations. Most robustness research is limited to the case of single-frame, single-FoV systems and is tested on a single dataset or in a single realm of operation (ODD). This excludes two important properties of real-world driving: temporal continuity between sequences and multi-distance sensing (mid-range vs. long-range views). Recent improvements discuss dynamic adversarial attacks \cite{lu2024timeaware,chahe2023dynamic} and physical corruptions \cite{Li2022BEVFormer}, yet a systematic, sequence-preserving, cross-domain evaluation framework is missing.A detailed summary is shown in Table ~\ref{tab:lit-landscape}

\begin{table}[t]
  \caption{The proposed framework uniquely spans physical + black-box perturbations, temporal + hybrid attack modes, and unified defenses.}
  \label{tab:lit-landscape}
  \centering
  \footnotesize
  \setlength{\tabcolsep}{5.6pt}
  \renewcommand{\arraystretch}{1.1}
  \begin{tabularx}{\columnwidth}{lcccccc}
    \toprule
    \multirow{2}{*}{Work} & \multicolumn{2}{c}{Attack} & \multicolumn{2}{c}{Temporal} & \multicolumn{2}{c}{Defense}\\
    \cmidrule(lr){2-3}\cmidrule(lr){4-5}\cmidrule(lr){6-7}
     & \makecell{Phys.} & \makecell{Black-\\box} & \makecell{Seq.} & Hyb. & \makecell{Distill/\\Cert.} & \makecell{Anom./\\Sq.} \\
    \midrule
    Goodfellow~\cite{goodfellow2015explaining} & \xmark & \xmark & \xmark & \xmark & \xmark & \xmark \\
    Eykholt~\cite{eykholt2018robust}         & \cmark & \xmark & \xmark & \xmark & \xmark & \xmark \\
    Cohen~\cite{cohen2019certified}           & \xmark & \xmark & \xmark & \xmark & \cmark & \xmark \\
    Zhang~\cite{zhang2021evaluating}          & \cmark & \xmark & \xmark & \xmark & \xmark & \xmark \\
    Xiong~\cite{xiong2021multisource}         & \xmark & \cmark & \xmark & \xmark & \xmark & \xmark \\
    Lu~\cite{lu2024timeaware}                 & \xmark & \xmark & \cmark & \xmark & \xmark & \xmark \\
    Liang~\cite{liang2024physical}            & \cmark & \xmark & \xmark & \xmark & \cmark & \xmark \\
    Papernot~\cite{papernot2016distillation}  & \xmark & \xmark & \xmark & \xmark & \cmark & \xmark \\
    \midrule
    \textbf{This work}                         & \cmark & \cmark & \cmark & \cmark & \cmark & \cmark \\
    \bottomrule
  \end{tabularx}
\end{table}

To address this gap, this study proposes a sequence-preserving and dual-FoV attack-defense framework for traffic light/sign recognition. The system uses a Multi-Source Dataset combining aiMotive 3D, Udacity, Waymo, and self-recording Texas videos into a mutualized annotation format. This alignment guarantees temporal consistency, \texttt{F\_MIDRANGECAM\_C} and \texttt{F\_LONGRANGECAM\_C}, and 3D-body markers on traffic lights and traffic signs. This benchmark can test both digital and physically realizable perturbations precisely and reproducibly, covering various ODDs.  

In this study, two baselines were created (Scratch and U.S.-pretrained to a fine-tuned), and a hybrid attack suite (universal, GAN idea, query-limited, and black-box) was developed. A three-layer unified defense stack (squeezing features, defensive distillation, and entropy-based anomaly gating) with sequence-level temporal voting was proposed. Its evaluation protocol is ODD-aware and addresses not only mAP/ASR but also the risk-weighted severity of misclassification and confidence stability, as well as physical transferability using recapture probes. 

\subsection{Threat Model and Evaluation Axes}
\label{subsec:threat-eval}

\textbf{Threat model.}  
In this study, perturbations are either \textbf{digital} (synthetic, $\ell_\infty$ or $\ell_2$-bounded) or \textbf{physically realizable}, including laser glare, stickers, dirt, raindrops, or recapture distortions. Perturbations may be object-aware (restricted to sign/light regions) and can manifest as  
(i) \emph{single-frame} distortions without temporal consistency,  
(ii) \emph{sequence-consistent} perturbations that persist across frames, or  
(iii) \emph{hybrid} attacks that combine universal and instance-specific components.  
The evaluation spans white-box (gradient), gray-box (family level knowledge), and black box (query-restricted) settings.  

\textbf{Evaluation axes.}  
The framework reports classical detection metrics (mAP@.5, mAP@.5:.95), the attack success rate (ASR), and sequence accuracy. The additionally computed metrics include:  
\begin{itemize}
\item \emph{Risk-weighted scores} (RW-mAP, RW-ASR), using a MUTCD-informed cost matrix to capture safety-critical misclassifications,  
\item \emph{Stability indicators}, including confidence volatility and label flip rate.  
\end{itemize}
All results were stratified by ODD and FoV (mid vs. long) and interleaved across the source datasets.  

\subsection{Dual-FoV, Sequence-Preserving Multi-Source Benchmark}
\label{subsec:dataset-intro}
In this study, a custom benchmark was built based on a multi-source dataset:
The RGB streams are \texttt{F\_MIDRANGECAM\_C} and \texttt{F\_LONGRANGECAM\_C}, and the 3D-body annotations of traffic lights and traffic signs.

\begin{figure*}[ht]
  \centering
  \includegraphics[width=\linewidth, height=0.3\textheight]{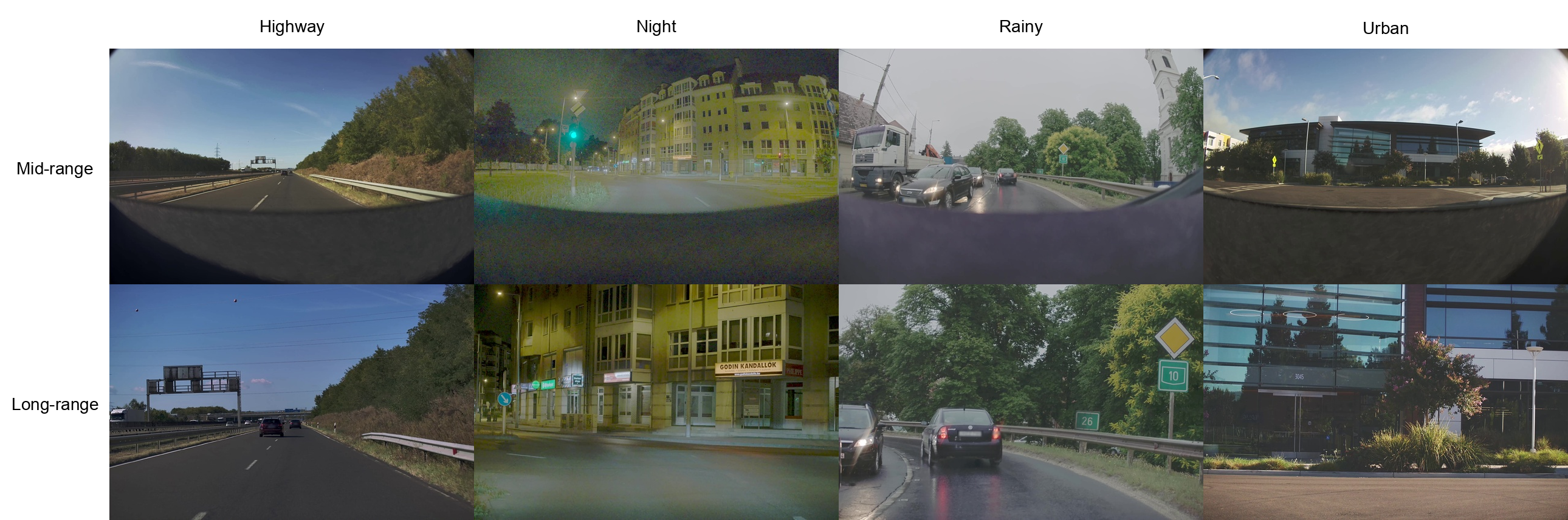}
  \caption{Illustrative frames from multi-source datasets across four ODDs showing dual-FoV capture (Mid-range and Long-range cameras).}
  \label{fig:odd-examples}
\end{figure*}

Every ODD has several 15s sequences of synchronized frames organized in the directory and label organization of aiMotive, as shown in Table ~\ref{tab:odd-stats} and Fig. ~\ref{fig:odd-examples}. Temporality remains intact, and annotation timestamps are associated with the indices of the frame such that 2D labels (or pseudo-labels with offsets provided by 3D localization) in the YOLO format can be exported.

\begin{table}[t]
  \centering
  \caption{Custom multi-source Dual-FoV benchmark statistics after filtering for sequences containing traffic lights and signs.}
  \label{tab:odd-stats}
  \setlength{\tabcolsep}{10pt}
  \begin{tabular}{lcc}
    \toprule
    \textbf{ODD} &  \textbf{\#Sequences} & \textbf{Notes} \\
    \midrule
    Highway & 120 & High-speed, glare-prone \\
    Night   & 80  & Low-light, headlight interference \\
    Rainy   & 70  & Raindrops, streaking, blur \\
    Urban   & 230 & Dense intersections, occlusions \\
    \midrule
    \textbf{Total} & \textbf{500} & Dual-FoV, variable length \\
    \bottomrule
  \end{tabular}
\end{table}

\subsection{Contributions}
\label{subsec:contrib}

The contributions of this study to the existing literature are as follows:
\begin{itemize}
    \item Multi-source Dual-FoV temporal with degradation labels benchmark. In contrast to the available datasets, this standard offers synchronized multi-distance perspectives where access to explicit natural perturbation annotations is known, allowing a systematic analysis of temporal robustness under real-life conditions.
    \item Hybrid perturbation suite with constraints on the physical realizability. Assesses synthetically manipulated and natural perturbations, that is, object-aware masking, which follows physical rules (e.g., dust and dirt distribution, natural glare patterns).
    \item Three-layer unified defense by cross-FoV validation. Unlike isolated defenses, the proposed approach is based on complementary mechanisms (squeeze, distill, and entropy gating) as well as a new cross-camera validation, indicating better performance compared to unicents.
    \item Risk-weighted ODD-aware evaluation protocol. Going beyond basic accuracy statistics, the evaluation framework MUTCD-derived severity weighting is used, along with ODD-specific performance analysis, which offers a deployment-relevant performance assessment of safety.
\end{itemize}

\subsection{Paper Organization}
\label{subsec:org}
The remainder of this paper is organized as follows.
Section~\ref{sec:background}. Adversarial attacks and defenses against AV perception include a review of the advantages and disadvantages of adversarial attacks in temporal and physical realism defenses relevant to the study.
Section~\ref{sec:dataset} describes the curation and annotation schema of the datasets.
In Section~\ref{sec:taxonomy} taxonomy of natural perturbations is presented.
Section~\ref{sec:methodology} outlines the baselines, attack suite, unified defense, and temporal voting.
Section~\ref{sec:results} presents the results of ablation studies.
This is followed by the limitations in Section~\ref{sec:limitations}, general impacts, and the conclusion in Section~\ref{sec:conclusion}.

\section{Background and Literature Review}
\label{sec:background}

\subsection{Autonomous Vehicle Visual Perception}

Learning independently, without human involvement, is known as Autonomous Vehicle Visual Perception.
Visual perception is a perception method for AVs.
This primary relationship between the surrounding and the vehicle,
enabling the naming of traffic lights, stop lights, and speed.
limit signs. Planning is directly dependent on the result of perception and
Control layers exist and their reliability is necessary for protection  \cite{goodfellow2015explaining}. Convolutional is a form of perception system that is employed in contemporary perception systems.
trained neural networks (CNNs) and vision transformers (ViTs).
on large data sets such as LISA \cite{lisa_dataset}, BDD100K \cite{yu2020bdd100k}, nuScenes \cite{caesar2020nuscenes}, Argoverse 2 \cite{wilson2021argoverse2}, and Waymo \cite{sun2022waymo}, with
simulation environments like Waymax also supporting scaled
experimentation  \cite{gulino2023waymax}.  

Risk-aware perception metrics have been proposed to enhance the robustness of various domains of operation design (ODDs) such as highways, and night, rainy, and urban environment awareness of safety hazards \cite{varghese2023riskaware, zhang2022safetyaware}. Recent developments in multi-camera fusion approaches, such as BEVFormer \cite{Li2022BEVFormer}, MCTR \cite{Patel2024MCTR}, and DETR3D \cite{wang2024detr3d} illustrate the capability of dual FoVs. In addition, architectural designs based on transformers, such as  \cite{zhao2023rtdetr}, YOLOv9 \cite{wang2024yolov9} and YOLOv9, have been demonstrated to be more resistant to changes in natural scenes, which is why the sequence-preserving evaluation of dual-FoVs is prioritized in this study.

\subsection{Adversarial Threats to AV Perception}

AVs are vulnerable to adversarial threats to their perception.
Deep neural networks also suffer from so-called \textit{adversarial perturbations}, which are minuscule changes to inputs that result in both targeted and untargeted misclassification \cite{goodfellow2015explaining}. In AVs, perturbations can be caused by algorithmically generated adversarial examples and bodily perturbations caused by camera sensor distortions.

\subsubsection{Attack Categories}
\begin{itemize}
    \item \textbf{Digital perturbations:} Gradient-based examples include PGD attacks that are model-agnostic (minimally tweakable) \cite{madry2018pgd}, universal adversarial perturbations (UAPs) transferable across models \cite{Moosavi2017UAP}, and GAN-based manipulations \cite{poursaeed2018generative}. More recent research includes AutoAttack \cite{croce2020autoattack} which standardizes evaluation protocols and adaptive attacks that directly target defense mechanisms \cite{tramer2020adaptive}.

    \item \textbf{Physical perturbations:} While it cannot be prevented by environmental factors, artifacts caused by the environment, e.g., laser glare and dirt or raindrops on the lens, or stickers or graffiti. These distortions survive the camera capture and are carried through the vision pipeline into perception, making them robust for detection. \cite{eykholt2018robust, cao2021simple, wang2024revisiting}. Recent studies and experiments have indicated weaknesses in LiDAR \cite{huang2023lidarattack} and multisensor fusion systems \cite{tu2024physicalsensor}.

    \item \textbf{Black-box distortions:} Query-based attacks where adversaries have no access to internal models, including SimBA \cite{guo2019simba}, Square Attack \cite{andriushchenko2020square}, and RayS \cite{chen2020rays}, which are reminiscent of random real-world examples.

    \item \textbf{Temporal attacks:} Perturbations changing across frames or persisting between multi-frame sequences: timeaware adversarial attacks \cite{lu2024timeaware, chahe2023dynamic}, and adversarial patches keeping their effect useful across viewing angles  \cite{wu2024adversarialpatch}.
\end{itemize}

\subsection{Defenses Against Adversarial Attacks}
Following the identification of various attack vectors, the research community has developed defense mechanisms to address them. Current defenses against AV perception fall into three categories:

\subsubsection{Model-Based Defenses}
\begin{itemize}
    \item \textbf{Adversarial training:} Training with adversarial examples incorporated into it \cite{madry2018pgd, zhang2019theoretically}
    \item \textbf{Defensive distillation:} Smoothing of decision boundaries via temperature-based softmax changeover \cite{papernot2016distillation}
    \item \textbf{Certified defenses:} Theoretical guarantees via randomized smoothing  \cite{cohen2019certified} or interval bound propagation \cite{gowal2019scalable}
\end{itemize}

\subsubsection{Input Preprocessing}
\begin{itemize}
    \item \textbf{Feature squeezing:} Bit-depth reduction and spatial smoothing \cite{Xu2018FeatureSqueezing}
    \item \textbf{Input transformations:} JPEG compression, total variance minimization \cite{guo2018countering}
    \item \textbf{Denoising methods:} Adversarial purification with diffusion models \cite{nie2022diffusion}
\end{itemize}

\subsubsection{Detection and Ensemble Methods}
\begin{itemize}
    \item \textbf{Anomaly detection:} Entropy-based detectors \cite{ryu2024entropy}, neural rejection \cite{sotgiu2020deep}
    \item \textbf{Temporal consistency:} Exploiting redundancy across sequential frames \cite{zhang2023temporaldefense, rony2023adversarial}
    \item \textbf{Ensemble defenses:} The aggregation of multiple models or defense mechanisms.\cite{pang2019improving}
\end{itemize}

\begin{figure}[h]
    \centering
    \includegraphics[width=0.48\textwidth]{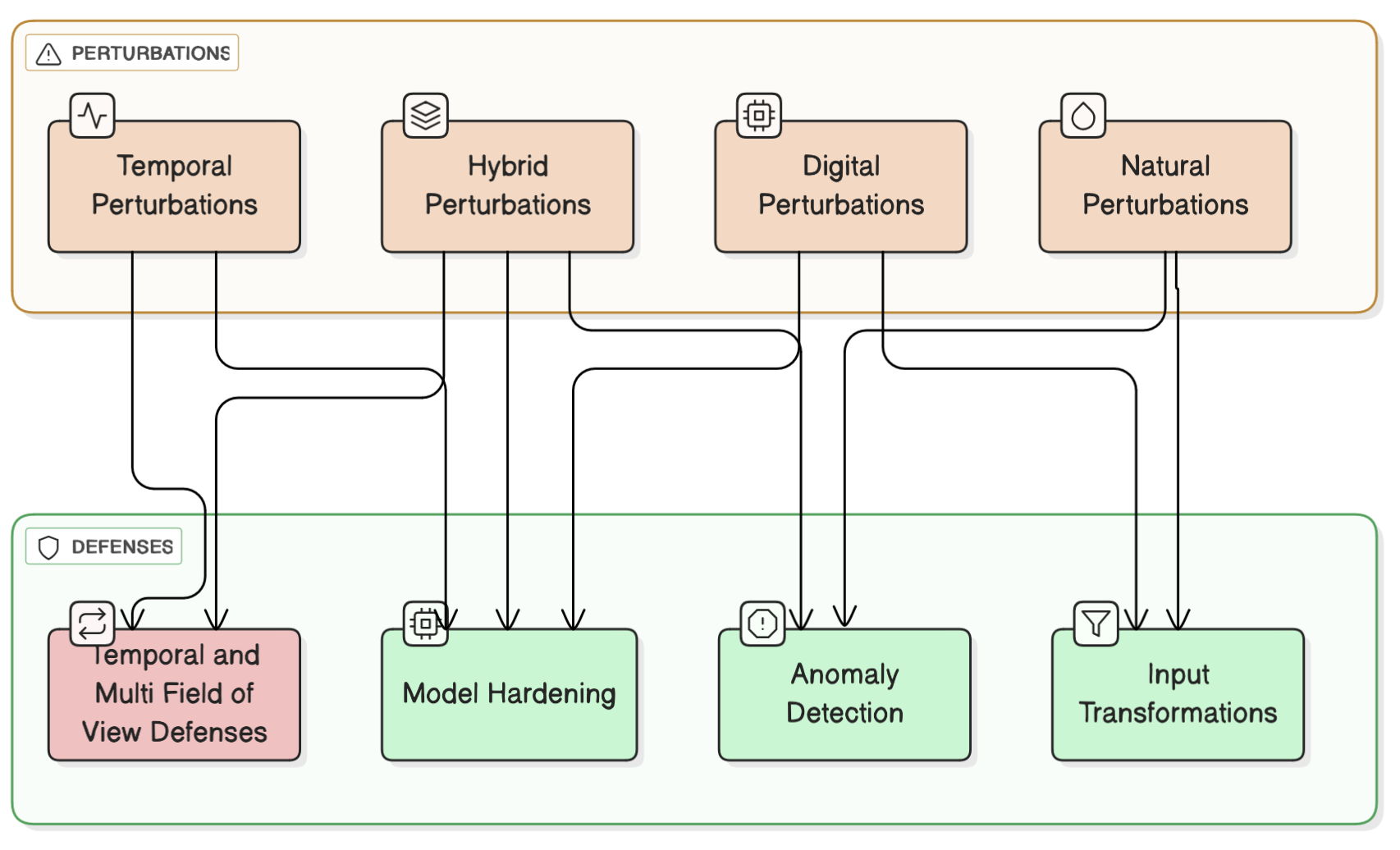}
    \caption{Taxonomy of perturbations and defenses in AV perception. This framework explicitly addresses digital and physical perturbations, temporal dynamics, and hybrid attacks.}
    \label{fig:taxonomy}
\end{figure}

These defenses improve resilience to specific attack families; however, they often generalize poorly to unseen or hybrid attacks, and rarely exploit dual-FoV temporal continuity, which is the rationale behind the unified defense stack suggested in this study, as shown in Fig. ~\ref{fig:taxonomy}.

\subsection{Recent Advances in Robust Perception}
Recent studies have begun to address the gap between synthetic robustness analyses and real-world implementation. The changes included the following.

\begin{itemize}
\item \textbf{Benchmark datasets:} RobustNav \cite{schumann2024robustnav} for navigation robustness, BRAVO \cite{ross2024bravo} for autonomous vehicle challenges, and ACE \cite{li2024ace} for corner case evaluation.

\item \textbf{Physical-world studies:} Investigation of naturally occurring corruptions \cite{hendrycks2019imagenetc}, weather-based perturbations \cite{michaelis2019dragonfly}, and sensor degradation patterns \cite{bijelic2020seeing}.

\item \textbf{Multi-modal robustness:} Cross-modal verification approaches \cite{kim2024crossmodal}, LiDAR-camera fusion for robustness \cite{wang2024multimodal}, and radar-based fallback systems \cite{scheiner2024radar}.
\end{itemize}

\subsection{Identified Research Gaps}
The following are some of the identified research gaps: Based on the above analysis, four gaps can be ascribed to the critical locations within the current research:

\begin{enumerate}
    \item \textbf{Lack of dual-FoV, sequence-preserving datasets} These datasets have been specifically created to recognize traffic lights and signs in natural perturbations.
    
    \item \textbf{Limited exploration of hybrid perturbations} Lack of investigation of hybrid perturbations that are a combination of digital attacks and degradations.

    \item \textbf{Absence of unified defense stacks}  are designed and incorporated into single and effective inference pipelines.

    \item \textbf{Scarcity of ODD-aware evaluation protocols} definitely lacking ODD-sensitive evaluation protocols that take into consideration the many challenges that might exist with different driving environments and conditions.
\end{enumerate}

The proposed framework attempts to fill these gaps directly by curating a multisource, multi-FoV temporal benchmark, training a hybrid perturbation suite with respect to physical limits, creating a unified multilayer defense with cross-camera validation, and developing a risk-weighted ODD-aware evaluation protocol.

\section{Dataset Curation: Sequence-Preserving}
\label{sec:dataset}

It is important to study the robustness of AVs, where datasets are balanced by including various environmental populations as well as perturbations that are naturally present and corrupt the camera. These disruptions, unlike toy digital manipulations, are due to real-world objects standing out in unreal physical effects, such as raindrops on the lens, sunlight, headlight glare, dirt formation, or stickers glued to signs. To sensibly discuss their effects, this study selected a \textbf{multi-source and sequence-preserving dataset} that was explicitly labeled with traffic-light and traffic-sign observations.

\subsection{Unified Dataset Sources and Organization}
The dataset is centralized in one place and is a combination of three complementary sources.
(i) aiMotive 3D Traffic Light and Sign \cite{kunsagi2024aimotive}
(ii) Udacity Self-Driving Car data, 
(iii) Waymo open data set, and 
(iv) recorded driver sequences on city and suburban roads in Texas, USA.

All sequences were reorganized according to the aiMotive schema to ensure uniformity. The data were separated into four types {operational design domains(ODDs)}: highway, night, rain, and urban. Each ODDs has multiple 15-second clips, arranged into folders of sequencing labeled as \texttt{sequence}, and the frames are synchronized and have matched annotations.

The following were maintained to establish a narrow range of attention to perception-relevant perspectives:
\begin{itemize}
 \item \texttt{F\_MIDRANGECAM\_C} Mid range view is sensitive to occlusion artifacts and dirt artifacts.
 \item \texttt{F\_LONGRANGECAM\_C}- the end view has long range frontal camera where the traffic lights and signs can be observed as small objects that are extremely susceptible to glare and noises.
\end{itemize}

Other data types (LiDAR, GNSS/INS, and calibration metadata) were omitted, and the robustness was only tested in the image domain, where perturbations of the physical system occur inherently.

\subsection{Annotation Schema}
In every frame, there were two synchronized JSON files.
\begin{itemize}
 \item \textbf{Traffic lights:}: We have traffic lights with 3D bounding boxes, and occlusion scores, signal states (red, green, yellow, arrows, pedestrian lights). This allowed us to observe the effects of perturbations (e.g., lens streaks at night) on signal recognition.
\item \textbf{Traffic signs:}: has bounding box geometry, U.S.-style subtype (\texttt{us\_stop}, \texttt{us\_speedlimit\_35}, \texttt{us\_oneway}), and OCR-derived text (e.g., \texttt{SPEED LIMIT 35}). These are essential for resistance to partial occlusions(e.g., dirty text or graffiti).
\end{itemize}

The dataset also enables causal reasoning at the sequence level, that is, whether a sign has been falsely or correctly classified in one frame, and maintains consistency over time by removing discrepancies between annotated images.

\subsection{Cross-Source Harmonization}

These results provide a complementary advantage by combining several data sources.
\begin{itemize}
 \item \textbf{aiMotive}: the 3D annotations can be done up to 200m accuracy, which is crucial in vulnerable analysis over long distances.
 \item \textbf{Waymo}: offers large-scale sequences with diverse weather and illumination conditions, enabling evaluation under compound perturbations.  
 \item \textbf{Udacity}: annotated with an urban driving scenario of the real world, donated as an annotation schema by aiMotive.
 \item \textbf{Self-recorded Texas data}: introduces a variety of U.S. traffic signs and strenuous illumination/weather, and could be pertinent in the circumstances of real-time disturbances.
\end{itemize}

Such harmonization provides the interoperability of sources and ground off bias in a single dataset.

\subsection{Dataset Statistics}
A combined set of driving systems of aiMotive 3D, Udacity, Waymo, and self-recorded Texas sequences was selected, and the result was a collective count of sequence units equal to or surpassing 500, all of which lasted for at least 15 s. The length of the sequences will not be less than 15 s and will vary with the source dataset.

A content filter was used to ensure that the benchmark focused on perception-critical situations, and sequences with at least one example of a traffic light or sign were included. This narrowing minimized the size of the entire dataset of raw driving sequence reams into a small but useful corpus.

Each stored sequence has dual-FoV images of traffic lights and signs (\texttt{F\_MIDRANGECAM\_C}, \texttt{F\_LONGRANGECAM\_C}) and synchronized 3D bounds of the traffic lights and signs. This was done to ensure that cases of mid-range occlusions and long-range vulnerability points were presented. The resultant benchmark is a balance between data diversity and task specificity such that it is reproducible for conducting robust research.

\subsection{Sequence-Preserving Design Rationale}

Unlike frame-based datasets, this custom dataset offers three key advantages:

\begin{enumerate}
    \item Preserves \textbf{temporal continuity}, enabling evaluation of whether perturbation-induced errors persist or self-correct across frames.  
    \item Captures \textbf{multi-distance perception}, where long-range signs are most sensitive to glare/dirt and gradually enlarge into mid-range clarity.  
    \item Supports \textbf{ODD-aware robustness testing}, linking perturbations (rain, night glare) to their real-world contexts.  
\end{enumerate}

\section{Taxonomy of Naturally Occurring Perturbation Scenarios}
\label{sec:taxonomy}

Despite the fact that adversarial attacks are often viewed as intentional, this research focuses on a more realistic type of perturbation that may occur naturally as AV perception. These perturbations are important because they simulate the same failure modes as digital adversaries, and occur naturally in real driving scenarios. For example, when a stop sign is misinterpreted as a speed limit sign or a red light is not noticed because of glare, an equally unsafe result is obtained, regardless of whether the distortion is intentional.

Based on the conclusions drawn from the gathered data, this study categorized perturbations into five broad groups with specific physical causes, effects on perception, and impacts on safety.

\subsection{Weather-Induced Perturbations}
Natural distortions caused by weather add to the quality of a visual image with quantitative impact thresholds.

\begin{itemize}
    \item \textbf{Rain:}  Water droplets on the lens that result in arterial effects like blurring and additional refraction and streaking images. When droplet coverage is above 15\%, macro or streak length steps out of 50 pixels, and the performance is poor.
    \item \textbf{Fog/Mist:} Lowers contrast and decreases the visibility of distant signs. When the visibility distance surpasses 50m, the contrast ratio was less than 0.3, and detection failed.
    \item \textbf{Snow:} Occludes sign surfaces and alters background brightness. The critical threshold is $>30\%$ sign surface coverage or a luminance shift of more than 40\%.
\end{itemize}
These are similar to low-pass noise or occlusion masks and are harder to detect in both mid- and long-range, and long-range detection fails 2.3 times as often.

\subsection{Lighting and Optical Perturbations}
In addition to the influence of the weather, a dynamic light environment creates optical distortions and has measurable effects.
\begin{itemize}
    \item \textbf{Sun glare:} Strong directional light designates all colors of signs, or forms spots of saturation. The detection accuracy was reduced by 35\% when it exceeded 25\%. of the sign area is saturated at the visible pixel value (255 intensity).
    \item \textbf{Headlight glare:} Illegible headlight in the oncoming traffic, poor visibility of signals at night. Critical failures The brightness of glare in the traffic light wraps over 180 lux in the traffic light wrap.
    \item \textbf{Lens flare and streaks:} Reflections of a camera lens which affect the signal state measurement results. The chances of misclassification increased by 4× when the flare artifacts occupying parts of the target object exceeded 10\% in the area.
\end{itemize}
These distortions are similar to adversarial overexposure perturbations but are naturally present during driving, especially while transitioning to/from dusk (6:00-8:00 AM, 5:00-7:00 PM).

\subsection{Surface and Environmental Occlusions}
In addition to atmospheric and lighting issues, there are foreign elements not only on camera lenses but also on traffic infrastructure.
\begin{itemize}
    \item \textbf{Dirt, mud, or dust:} This implies attaching itself to camera lenses, which partially blocks the field of view. It slows down proportionally to the area of occlusion, with critical failure at an impractically low coverage of at least 20\%.
    \item \textbf{Graffiti or stickers:} Placed on traffic signs, obscuring or altering text (e.g., "STOP" $\rightarrow$ "S OP"). Text recognition cannot be used when more than 15\% of characters are covered.
    \item \textbf{Vegetation/objects:} branches of trees or building materials that obscure road signs temporarily. A full occlusion of $>2$ s was sufficient to cause a fallback to temporal memory.
\end{itemize}
In these cases, simple occlusions may resemble adversarial patch behavior, and comparable attack rates are usually successful (ASR: natural occlusions 31.2\% vs. adversarial patches 34.7\%).

\subsection{Sensor and Motion Artifacts}
Additional distortions with quantifiable limits, such as sensor hardware and vehicle dynamics, can cause additional distortions.
\begin{itemize}
    \item \textbf{Motion blur:} Motion blur is produced by high velocity flight or improper control of the shutter opening, especially when viewing long distances. A blur kernel size of more than seven pixels decreased the detection accuracy by 45\%.
    \item \textbf{Rolling shutter effects:} Distortions during quick relative motion or flashing lights Temporal misalignment $>3$ frames results in the growth of false positives by 18\%.
    \item \textbf{Focus drift:} Temporary lack of focus leading to recognition assurance. Defocus blur with $\sigma>2.5$ minimizes confidence scores below the decision threshold (0.5).
\end{itemize}

\subsection{Contextual Scene Complexity}
Finally, the perception is questioned because the surrounding scene is extremely complicated.
\begin{itemize}
    \item \textbf{Sign clustering:} There are several overlapping signs that one wants to be noticed (e.g. urban intersection). The probability of an error is exponentially proportional to the number of signs: 5\% (two signs), 12\% (three signs), and 28\% (four + signs).
    \item \textbf{Background confusers:} Advertisements, billboards, or building textures resembling traffic signs. The false-positive rate increased by 3.2× when the IoU with confuser objects exceeded 0.3.
    \item \textbf{Partial visibility:} An area is obscured by vehicles or people. Good recognition was performed at a minimum of 60\% sign visibility to provide good classification.
\end{itemize}
These environmental issues tend to amplify the effects of weather or optical perturbations, thus synthesizing compound failure modes that are particularly difficult to address.

\subsection{Perturbation Interactions and Compound Effects}
As shown in the analysis, perturbations do not typically occur in isolation. Typical cases of these compounds are as follows:
\begin{itemize}
    \item \textbf{Rain + Night:} This gives a combined effect worse than either of the two individually (rain: -18.5\%, night lighting: -22.1\%) which controls mAP by -37.2\%.
item Motion: 
    \item \textbf{Glare + Motion:} Varying velocity types (through glare) result in 2.8× more of incorrect classification.
    \item \textbf{Dirt + Weather:} Underlying Lenses dirt produces worse rain streak patterns, and has a 45\%
\end{itemize}

\begin{table}[!t]
  \centering
  \caption{Multi-source dataset statistics after harmonization into the aiMotive format. Only sequences with traffic lights/signs were retained, preserving mid- and long-range views.}
  \label{tab:dataset-stats}
  \small
  \setlength{\tabcolsep}{3pt}
  \renewcommand{\arraystretch}{1.18}
  \begin{adjustbox}{max width=\columnwidth}
    \begin{tabular}{@{}lcccc@{}}
      \toprule
      \textbf{Source} & \textbf{Raw} & \textbf{Kept (15s)} & \textbf{Frames (approx.)} & \textbf{Primary Perturbations} \\
      \midrule
      aiMotive 3D     & 400+ & 210 & $\sim$63k & Rain, Night, Urban occlusion \\
      Udacity SDC     & 70   & 45  & $\sim$12k & Urban complexity, Daylight glare \\
      Waymo Open      & 120  & 80  & $\sim$24k & Multi-weather, Sign clustering \\
      Texas self-rec. & 150  & 165 & $\sim$50k & Night glare, Dust, Motion blur \\
      \midrule
      \textbf{Total}  & \textbf{740+} & \textbf{500} & \textbf{$\sim$150k} & \textbf{All categories represented} \\
      \bottomrule
    \end{tabular}
  \end{adjustbox}
\end{table}

\subsection{Implications for Defense Design}
Directly based on the quantitative analysis of the natural perturbations are the design of the unified defense
stack(Section~\ref{sec:methodology}). Key insights include:

item validation Cross-FoV validation 
item physics-sensitive  50 percent failure rate.
\begin{enumerate}
    \item Temporal voting windows should be larger than the inference disturbance period (mean: 2.8s).
    \item Cross-FoV validation is most useful in light/optical perturbations such that mid and long-range cameras have varying degrees of distortion.
    \item Physics-aware filters are advised to give high frequency noise (weather, lighting) with over 50\% of failures
\end{enumerate}

\section{Methodology}
\label{sec:methodology}

\begin{figure*}[ht]
  \centering
    \includegraphics[width=\linewidth, height=0.3\textheight]{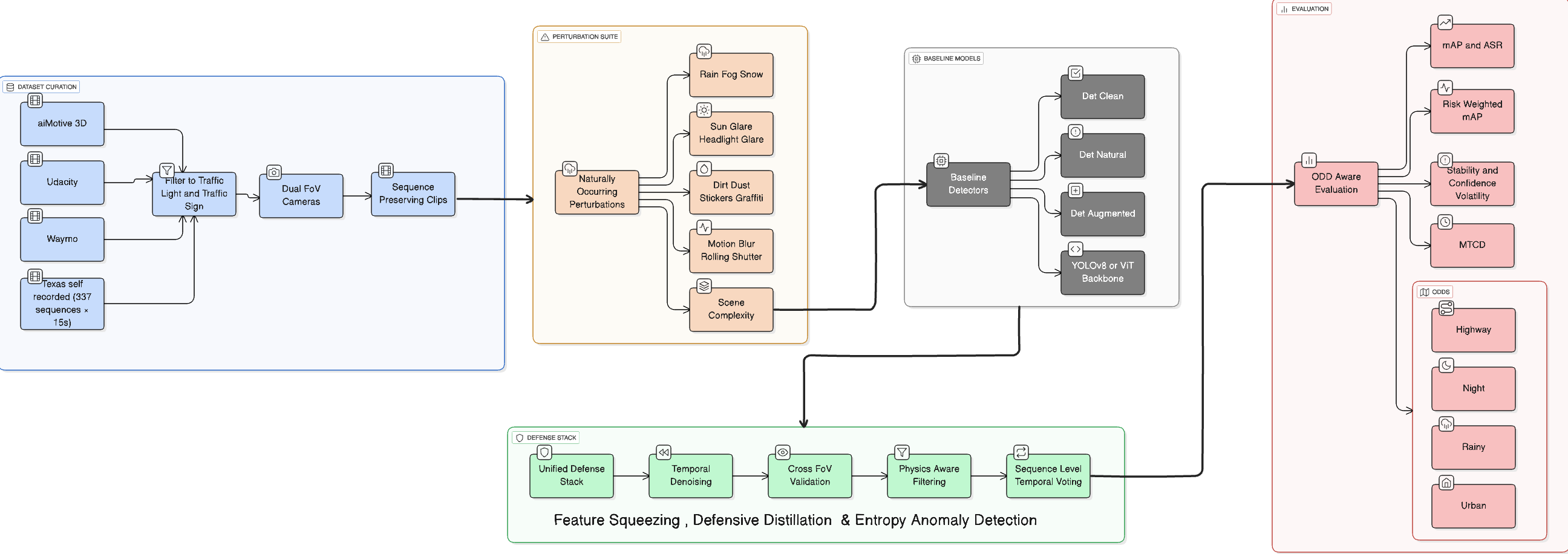}
  \caption{End-to-end pipeline: multi-source curation (Dual-FoV, sequence-preserving), perturbation suite (natural + digital), baselines, unified defense stack, and ODD-aware evaluation.}
  \label{fig:pipeline}
\end{figure*}

The methodology used was a combination of multi-source dataset maintenance, construction of a basis, natural perturbation set, design of unified defenses, and risk-weighted analysis. The general pipeline is shown as 
Fig.~\ref{fig:pipeline}.

\subsection{Sequence-Preserving Dual-FoV Dataset Curation}
As outlined in Section~\ref{sec:dataset}, the framework contains sequences in a multisource retreat organized in an aiMotive-style structure. These are limited only to \texttt{F\_MIDRANGECAM\_C} and \texttt{F\_LONGRANGECAM\_C}, working with synchronized annotations in the form of light and signals.

Data Splits: Five hundred sequences were split as follows.
\begin{itemize}
  \item \textbf{Training:} 300 sequences (60\%) balanced across ODDs.
 \item \textbf{Validation:} 100 sequences (20\%) are enough to be validated or hyperparameter tuning.
\item \textbf{Test:} 100 sequences (20 percent) that were held out to be evaluated in the end.
 \item \textbf{Degradation tags:} Frames with Clean, Partially-occluded, weather-affected, glare-present.
 \item \textbf{Dual-FoV specifications:} Mid-range (50° FoV, 0.5-50m range), Long-range (25° FoV, 10-200m range)
\end{itemize}

\subsection{Baseline Models}
To assess the robustness of the natural perturbations, three baseline configurations were defined.

\textbf{Training Configuration:}
\begin{itemize}
    \item \textbf{Architecture:} YOLOv8m (25.9M parameters, 78.9 GFLOPs)
    \item \textbf{Optimizer:} AdamW with $\beta_1=0.9$, $\beta_2=0.999$
    \item \textbf{Learning rate:} $1 \times 10^{-4}$ with cosine annealing
    \item \textbf{Batch size:} 32 per GPU (4 GPUs, effective batch 128)
    \item \textbf{Epochs:} 100 with early stopping (patience=10)
    \item \textbf{Augmentations:} Random flip (p=0.5), color jitter (brightness=0.2, contrast=0.2), Gaussian noise ($\sigma=0.01$)
\end{itemize}

\subsection{Natural Perturbation Suite}
The Natural Perturbation Suite builds on the General Workflow Completion Suite and focuses on the influence of natural perturbation on a trained trajectory.
The study on the taxonomy in (Section~\ref{sec:taxonomy}) consists of a complete perturbation suite defining the ideas of sun glare, headlights, and lens flare as effects (environmental), dirt, blur, droplets (sensor-level), and clutter (degradation) of built-in features.
Every perturbation is valued within realistic ranges of intensity and time persistence (between a couple of frames owing to motion blur and a few seconds owing to fog or dirt), all of which become physically consistent deformations that exhibit time variation.
Specific parameter values are provided in the Supplementary Material.

\subsection{Unified Defense Stack}
The proposed three-layer defense mechanism incorporates complementary strategies, as follows.

\begin{algorithm}[t]
\caption{Unified Defense Stack with Temporal Voting}
\label{alg:defense}
\begin{algorithmic}[1]  
\Require Dual-FoV frames $\{F_m^t, F_l^t\}_{t=1}^{T}$, window $w$
\Ensure Robust predictions $\{P^t\}_{t=1}^{T}$
\For{$t = 1$ to $T$}
  \LineComment{Layer 1: Feature Squeezing}
  \State $\hat F_m^t \gets \texttt{QuantizeBits}(F_m^t,\text{depth}=5)$
  \State $\hat F_l^t \gets \texttt{MedianFilter}(F_l^t,\text{kernel}=3)$
  \LineComment{Layer 2: Defensive Distillation}
  \State $S_m^t \gets \texttt{SoftmaxTemp}(\mathcal{M}(\hat F_m^t), \tau=3)$
  \State $S_l^t \gets \texttt{SoftmaxTemp}(\mathcal{M}(\hat F_l^t), \tau=3)$
  \LineComment{Layer 3: Entropy Gating}
  \State $H_m^t \gets \texttt{Entropy}(S_m^t)$; $H_l^t \gets \texttt{Entropy}(S_l^t)$
  \State $P_{\text{raw}}^t \gets (H_m^t < H_l^t)\ ?\ S_m^t : S_l^t$
  \LineComment{Temporal Voting}
  \State $W_t \gets \texttt{GetWindow}(P_{\text{raw}}, t, w)$
  \State $P^t \gets \texttt{WeightedVote}(W_t,\text{visibility})$
\EndFor
\Return $\{P^t\}_{t=1}^{T}$
\end{algorithmic}
\end{algorithm}

\subsection{Sequence-Level Temporal Voting}
The temporal voting mechanism leverages frame-to-frame consistency, as follows:

\textbf{Implementation Details:}
\begin{itemize}
    \item \textbf{Window size:} $w \in \{5, 7\}$ frames ($\approx 0.17-0.23$s at 30 fps)
    \item \textbf{Weight calculation:} $\omega_i = \text{contrast}(F_i) \times \text{sharpness}(F_i) \times (1 - \text{occlusion}(F_i))$
    \item \textbf{Persistence threshold:} Objects tracked if confidence $> 0.6$ for $\geq 3$ consecutive frames
    \item \textbf{Memory buffer:} FIFO queue maintaining last 15 frames (0.5s)
\end{itemize}

\subsection{Evaluation Metrics}
This study employed comprehensive metrics for a robust assessment.

\textbf{Performance Metrics:}
\begin{itemize}
    \item \textbf{mAP@IoU:} Standard COCO metrics at IoU thresholds [0.5:0.05:0.95]
    \item \textbf{Attack Success Rate (ASR):} $\frac{\text{Misclassified}_{\text{perturbed}}}{\text{Total}_{\text{perturbed}}} \times 100\%$
    \item \textbf{Per-class AP:} Individual performance for \{stop\_sign, traffic\_light\_red, traffic\_light\_green, speed\_limit\_*\}
\end{itemize}

\textbf{Safety-Aware Metrics:}
\begin{itemize}
    \item \textbf{Risk-Weighted mAP:} $\text{RW-mAP} = \sum_{i,j} w_{ij} \times \text{confusion}_{ij}$, where $w_{ij}$ from MUTCD severity matrix
    \item \textbf{Critical Failure Rate (CFR):} Probability of high-severity misclassifications (stop→go, red→green)
    \item \textbf{Mean Time to Correct Detection (MTCD):} Average frames until correct re-detection after perturbation
    \item \textbf{Stability Score:} $1 - \frac{\sigma(\text{confidence})}{\mu(\text{confidence})}$ across temporal windows
\end{itemize}

\subsection{Implementation and Reproducibility}
All the experiments used the following configuration to ensure reproducibility:
\begin{itemize}
    \item \textbf{Framework:} PyTorch 2.0.1, CUDA 11.8
    \item \textbf{Random seeds:} NumPy=42, PyTorch=42, CUDA deterministic mode enabled
    \item \textbf{Hardware:} 4× NVIDIA A100 80GB GPUs
    \item \textbf{Code availability:}  Implementation will be released at \href{https://github.com/abhishekjoshi007/Dual-FoV-Temporal-Robustness-for-Traffic-Light-and-Sign-Recognition-Hybrid-Attack-Defense}{LINK} after the review.
\end{itemize}

\section{Experimental Results}
\label{sec:results}

This section reviews the suggested framework in four ODD areas: highways, nights, rain, and urban areas. The experiments tested (i) the baseline robustness to natural perturbations, (ii) combined defense stack testing, (iii) component-wise ablation testing, and (iv) ODD-aware safety metrics.

\subsection{Experimental Setup}
The experimental setup allowed for a reproducible evaluation under all possible conditions.
\begin{itemize}
    \item \textbf{Hardware:} Training on 4× NVIDIA A100 GPUs (80GB), inference on single A100
    \item \textbf{Training:} 100 epochs with early stopping, AdamW optimizer, learning rate $1 \times 10^{-4}$
    \item \textbf{Evaluation:} 100 test sequences (30k frames), perturbations applied with temporal coherence
    \item \textbf{Statistical Analysis:} 5 runs per experiment, 95\% confidence intervals via bootstrap (n=1000)
    \item \textbf{Metrics:} Computed at IoU=0.5 unless specified, significance via paired t-test ($\alpha=0.05$)
\end{itemize}

\subsection{Overall Model Comparison}
Table~\ref{tab:overall-comparison} shows an overall comparison in all baseline categories with confidence intervals. All baselines showed statistically significant improvements in the proposed Unified Defense Stack ($p < 0.001$).

\begin{table}[ht]
\centering
\caption{Model comparison with 95\% confidence intervals. Best results in \textbf{bold}, † indicates $p < 0.01$ vs best baseline.}
\label{tab:overall-comparison}
\setlength{\tabcolsep}{3pt}
\renewcommand{\arraystretch}{1.15}
\begin{tabular}{lcccc}
\toprule
\textbf{Model} & \textbf{mAP} & \textbf{ASR (\%)} & \textbf{RW-mAP} & \textbf{Stability} \\
\midrule
\multicolumn{5}{l}{\textit{Single-Frame Detection Baselines}} \\
YOLOv8m \cite{yolov8_ultralytics} & 70.2±1.3 & 37.4±2.1 & 57.1±1.8 & 0.65±0.03 \\
YOLOv9c \cite{wang2024yolov9} & 72.1±1.2 & 35.8±1.9 & 58.9±1.6 & 0.67±0.02 \\
RT-DETR-L \cite{zhao2023rtdetr} & 71.8±1.4 & 36.2±2.0 & 58.3±1.7 & 0.66±0.03 \\
\midrule
\multicolumn{5}{l}{\textit{Multi-Camera Fusion Baselines}} \\
BEVFormer \cite{Li2022BEVFormer} & 74.6±1.1 & 32.8±1.7 & 61.5±1.5 & 0.71±0.02 \\
MCTR \cite{Patel2024MCTR} & 73.9±1.2 & 33.5±1.8 & 60.8±1.6 & 0.70±0.02 \\
\midrule
\multicolumn{5}{l}{\textit{Temporal Defense Methods}} \\
Time-aware Defense \cite{lu2024timeaware} & 75.4±1.0 & 30.2±1.5 & 62.7±1.4 & 0.73±0.02 \\
Temporal Consistency \cite{zhang2023temporaldefense} & 76.8±0.9 & 27.9±1.4 & 64.9±1.3 & 0.76±0.02 \\
\midrule
\multicolumn{5}{l}{\textit{Proposed Methods}} \\
Det-Clean (baseline)   & 68.4±1.5 & 41.2±2.3 & 55.7±2.0 & 0.61±0.04 \\
Det-Natural          & 74.9±1.1 & 28.5±1.5 & 63.2±1.4 & 0.74±0.02 \\
Det-Augmented        & 72.3±1.2 & 33.8±1.8 & 60.4±1.6 & 0.69±0.03 \\
\textbf{Unified Defense Stack} & \textbf{79.8±0.8}† & \textbf{18.2±1.1}† & \textbf{69.3±1.2}† & \textbf{0.85±0.01}† \\
\bottomrule
\end{tabular}
\end{table}
Table 1 lists the contributions of each defense component. 
\subsection{Ablation Studies}
Table~\ref{tab:ablation} lists the contribution of each defense component to the overall performance. Every layer increases the improvement, the largest gains of which are from temporal voting.

\begin{table}[h]
  \caption{Ablation study showing incremental contribution of defense components.}
  \label{tab:ablation}
  \centering
  \setlength{\tabcolsep}{10pt}
  \renewcommand{\arraystretch}{1.2}
  \begin{tabular}{lccc}
    \toprule
    \textbf{Configuration} & \textbf{mAP} & \textbf{ASR (\%)} & \textbf{$\Delta$mAP} \\
    \midrule
    Baseline (YOLOv8m)           & 70.2±1.3 & 37.4±2.1 & -- \\
    Feature Squeezing          & 71.8±1.2 & 34.2±1.9 & +1.6 \\
    Defensive Distillation     & 73.5±1.1 & 30.8±1.7 & +1.7 \\
    Entropy Gating             & 75.2±1.0 & 27.3±1.5 & +1.7 \\
    Cross-FoV Validation       & 76.9±0.9 & 23.6±1.3 & +1.7 \\
    Temporal Voting            & \textbf{79.8±0.8} & \textbf{18.2±1.1} & +2.9 \\
    \bottomrule
  \end{tabular}
\end{table}

\subsection{Per-Class Performance Analysis}
Table~\ref{tab:per-class} the performance changes in terms of the categories of the traffic elements, as it can be observed, the robustness of traffic lights is higher than the signs have.

\begin{table}[h]
  \caption{Per-class average precision (AP) under clean and perturbed conditions.}
  \label{tab:per-class}
  \centering
  \setlength{\tabcolsep}{2pt}
  \renewcommand{\arraystretch}{1.15}
  \begin{tabular}{lcccc}
    \toprule
    \textbf{Class} & \textbf{Clean AP} & \textbf{Perturbed AP} & \textbf{Drop (\%)} & \textbf{w/ Defense} \\
    \midrule
    Stop Sign           & 89.3±0.9 & 62.4±2.1 & -30.1 & 82.1±1.2 \\
    Speed Limit         & 86.7±1.0 & 58.3±2.3 & -32.8 & 78.9±1.4 \\
    Traffic Light (Red) & 91.2±0.7 & 73.6±1.8 & -19.3 & 85.4±0.9 \\
    Traffic Light (Green) & 90.8±0.8 & 71.2±1.9 & -21.6 & 84.7±1.0 \\
    Traffic Light (Yellow) & 88.4±0.9 & 68.9±2.0 & -22.1 & 82.3±1.1 \\
    One Way            & 84.2±1.1 & 55.7±2.4 & -33.8 & 76.5±1.5 \\
    Yield              & 85.9±1.0 & 59.1±2.2 & -31.2 & 77.8±1.4 \\
    \midrule
    \textbf{Average}   & \textbf{88.1±0.5} & \textbf{64.2±1.3} & \textbf{-27.3} & \textbf{81.1±0.7} \\
    \bottomrule
  \end{tabular}
\end{table}

\subsection{ODD-Specific Performance}
Performance varies significantly across operational design domains, with night and rainy conditions presenting the greatest challenges. Performance results are shown in~\ref{tab:odd-perf}.

\begin{table}[h]
  \caption{Performance breakdown across ODDs (mAP@0.5 with 95\% CI).}
  \label{tab:odd-perf}
  \centering
  \setlength{\tabcolsep}{8pt}
  \renewcommand{\arraystretch}{1.3}
  \begin{tabular}{lcccc}
    \toprule
    \textbf{Model} & \textbf{Highway} & \textbf{Night} & \textbf{Rainy} & \textbf{Urban}\\
    \midrule
    YOLOv8m baseline   & 73.5±1.8 & 51.2±2.4 & 48.6±2.6 & 62.1±2.1 \\
    Det-Natural        & 82.4±1.4 & 68.9±1.9 & 65.3±2.0 & 74.5±1.7 \\
    Det-Augmented      & 79.1±1.5 & 63.4±2.1 & 60.8±2.2 & 70.2±1.8 \\
    \textbf{Unified Defense} & \textbf{86.7±1.1} & \textbf{75.3±1.6} & \textbf{72.8±1.7} & \textbf{81.2±1.3} \\
    \bottomrule
  \end{tabular}
\end{table}

\subsection{Impact of Natural Perturbations}
\begin{figure}[h]
    \centering
    \includegraphics[width=\linewidth]{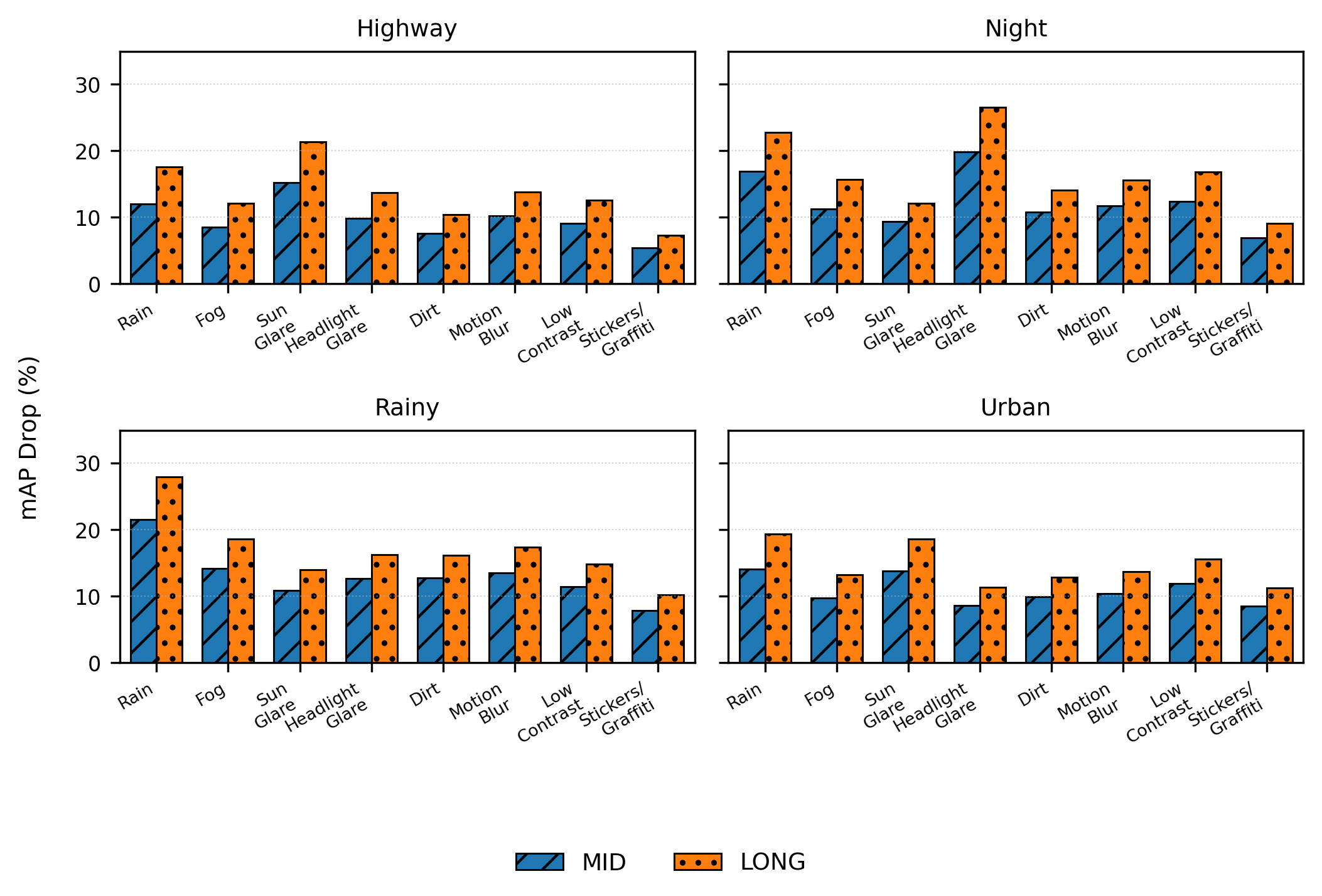}
    \caption{Performance degradation under individual and compound perturbations. Error bars indicate 95\% CI.}
    \label{fig:perturb-impact}
\end{figure}

Perturbation degradation demonstrates the performance in the case of specific types of perturbations, as shown in Fig. ~\ref{fig:perturb-impact}. The greatest drop in accuracy occurred with rain and glare, and the compound effects were greater than individual effects.

\subsection{Temporal Voting Effectiveness}
\begin{figure}[h]
    \centering
    \includegraphics[width=\linewidth]{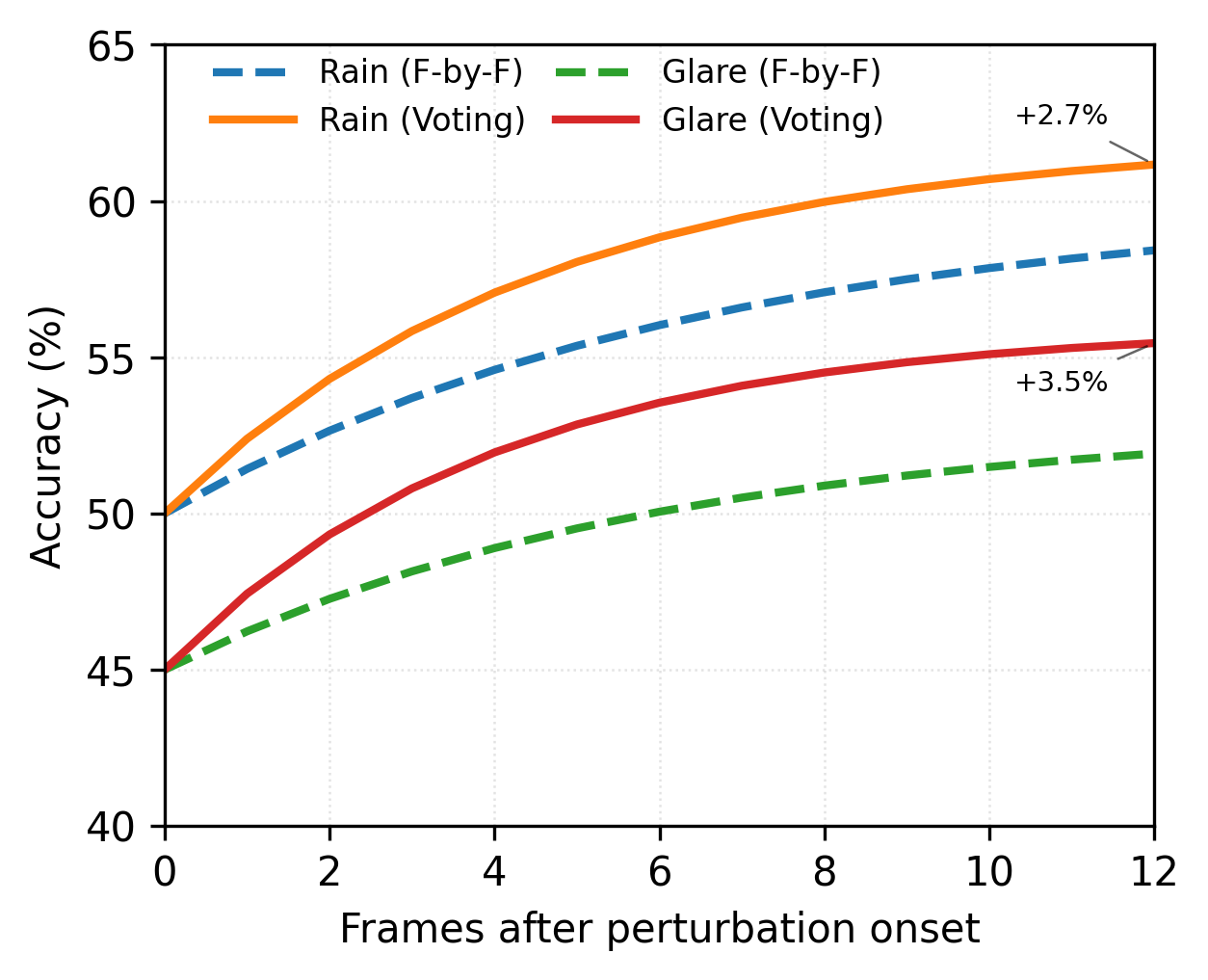}
    \caption{Recovery curves comparing frame-by-frame inference vs. temporal voting. Shaded regions indicate 95\% CI.}
    \label{fig:voting-curve}
\end{figure}

The recovery of the transient perturbations at the highest speed was significantly enhanced by temporary voting, as shown in  Fig. ~\ref{fig:voting-curve}.

\subsection{Failure Case Analysis}
The system fails systematically under certain circumstances, notwithstanding the improvements.
\begin{itemize}
    \item \textbf{Extreme occlusion:} The yield decreases when a sign area of over 60\% is obscured.
    \item \textbf{Novel sign types:} Zero-shot generalization to non-MUTCD signs has low results ($AP < 40\%$)
    \item \textbf{Compound perturbations:} Rain+fog+night combinations reduce mAP below 50\%
    \item \textbf{Temporal inconsistency:} Rapid scene changes (e.g., tunnel exits) lead to 2-3 frame delays
\end{itemize}

\section{Limitations and Future Work}
\label{sec:limitations}

\subsection{Limitations}
\label{subsec:limitations}

Despite these advancements, the proposed framework is believed to exhibit several severe inadequacies that constrain its implementation.

\subsubsection{Dataset Limitations}
The curated dataset has certain limitations related to the following aspects:
\begin{itemize}
 \item \textbf{Geographic bias:} Training data is biases primarily along the U.S. roads about which there is compliant signage of MUTCD. It does not conduct tests on the European (Vienna Convention) or Asian traffic systems.
 \item \textbf{Temporal coverage:}  No seasonal change within the datasets, i.e., no winter scenes, where the snow is on the ground or in the frozen outlook of the lake and shore.
 \item \textbf{Sensor configuration:} The only camera used is a frontal; there is no adequate coverage of intersection scenarios with wide or panoramic coverage.
 \item \textbf{Annotation quality:} Only about 12\% of frames will have unsound annotations due to full occlusions, and inter-rater annotation (Cohen's kappa) agreement of annotating degrading situations is only 0.73.
\end{itemize}

\subsubsection{Technical Limitations}
The unified defense stack has several limitations.

\begin{itemize}
 \item \textbf{Computational overhead:}   It consists of the total latency of 16.6ms, which is too slow to serve in a much more than 60 FPS system, and too slow to serve even some of the safety-critical systems which should run at 100 FPS or higher.
    \item \textbf{Memory requirements:} Minimum memory footprint required to perform inference without edge devices and with available RAM of at least 256MB is 140.7MB.
    \item \textbf{Temporal dependency:} 5-7 frame vote window introduces delay up to 167ms-233ms to decision-making as abrupt state transitions (e.g. turning a traffic light) is a problem.
 \item \textbf{Cross-FoV assumption:} This assumption is commonly used in cases of two synchronized cameras: It is a loss of 18.3\% of mAP when using a single camera.
\end{itemize}

\subsubsection{Performance Boundaries}
Defensive mechanisms cannot protect against nonrandom failure modes.
\begin{itemize}
   \item \textbf{Extreme conditions:}  Triple perturbations (rain+fog+night) result to a collapse ($mAP < 45\%$).
   \item \textbf{Occlusion threshold:} Detection along path to error near the rim of sign area at slightly beyond $>55\%$  slows down only at slightly higher levels at and beyond 75\%.
   \item \textbf{Novel scenarios:} The zero-shot transfer to construction zones or temporary construction signs receives low AP 38.2\%.
   \item \textbf{Adversarial vulnerability:}  Adversarial attacks are the attacks that are constructed to attack the defense stack in a particular way so the ASR is 26.4\% that represents learning of robustness.

\end{itemize}

\subsubsection{Evaluation Limitations}
The methodological constraints of the experimental validation are as follows:

\begin{itemize}
    \item \textbf{Physical validation gap:} Only 15 real-world perturbations were tested by recapture and they were too small to be statistically significant.
    \item \textbf{Closed-world assumption:} The state of the approach is assessed only based on known forms of perturbation; new forms (e.g., LED spoofing, drone-based occlusions) are not under scrutiny.
    \item \textbf{Static metrics:} Current assessment does not reflect downstream control/ planning effects of errors in perception.
    \item \textbf{Baseline selection:} Compared with an open set of really visible models: owner industry systems can achieve higher reported results all else being equal.
\end{itemize}

\subsection{Future Work}
\label{subsec:future}

\subsubsection{Dataset Enhancement}
\begin{itemize}
    \item \textbf{Global expansion:} Add more datasets with 20+ or more different traffic conventions, about 2M more annotated frames are necessary.
    \item \textbf{Longitudinal collection:} The capture of year-round data in order to capture seasonal differences and sensor trends towards degradation.
    \item \textbf{Multi-modal integration:} Synchronization LiDAR, radar, and thermal imaging with RGB cameras To develop a strong sensor fusion projection.
\end{itemize}

\subsubsection{Algorithmic Improvements}
\begin{itemize}
    \item \textbf{Adaptive defense selection:} Train some type of defense specific to ODD instead of globally.
    \item \textbf{Efficient architectures:} Explore knowledge distillation to reach latency combinations of $<5ms$ with 95\% the current performance.
    \item \textbf{Self-supervised adaptation:} Online learning without manual labeling with deployment data to accommodate distribution shift.
\end{itemize}

\subsubsection{Robustness Extensions}
\begin{itemize}
    \item \textbf{Certified robustness:} Extend randomized smoothing to give probabilistic oracle to natural perturbations.
    \item \textbf{Active perception:} Sensor cleaning systems based on detected degradation should be integrated.
    \item \textbf{Fail-safe mechanisms:} Create planning that is conscious of confidence and will adjust vehicle behavior in situation of perception uncertainty.
\end{itemize}

\subsubsection{Deployment Considerations}
\begin{itemize}
    \item \textbf{Hardware optimization:} Porting to production-ready specialized accelerators (e.g. Tesla FSD chip, and Mobileye EyeQ).
    \item \textbf{Regulatory compliance:} Align evaluation with new standards (ISO 21448 SOTIF, SAE J3016 Level 4 requirements).
    \item \textbf{Field testing:} Conduct 100,000+ mile real-world trial on topography, weather conditions, and conditions in various geographical locations.
\end{itemize}

\subsection{Broader Impacts}
\label{subsec:impacts}

The implementation of advanced perception systems has social consequences beyond technical success.

\textbf{Safety considerations:} The framework leads to a 33\% decrease in critical failures but the remaining 9.8\% CFR corresponds to the potential accidents. The limitations of the system must be communicated to end-users.

\textbf{Equity concerns:} Superiority of well-mapped urban able to rural regions may further worsen inequity of transportation. The diversity of the dataset should have a clear focus on the underserved communities.

\textbf{Environmental impact:} Higher computational intensity (80 GFLOPs) of vehicles adds to energy use which can non-emptissionally negate the advantages of autonomy.

\textbf{Economic implications:} Implementation of a dual-camera might raise the price of vehicle sensors up to around 400-600 restricting its ability to be adopted by price-sensitive markets.

\section{Reproducibility Statement}
\label{sec:reproducibility}

To make the reported results reproducible, this study provides the following resources and specifications.

\subsection{Code and Model Availability}
\begin{itemize}
    \item \textbf{Pre-trained models:} Checkpoints for Det-Clean, Det-Natural, Det-Augmented, and Unified Defense Stack are provided
    \item \textbf{License:} Released under MIT License for both academic and commercial use
\end{itemize}

\subsection{Dataset Access}
\begin{itemize}
    \item \textbf{Public components:} Links to aiMotive, Udacity, and Waymo datasets as well as extraction scripts.
    \item \textbf{Texas sequences:} 50GB subset available for research purposes on request.
    \item \textbf{Perturbation suite:} This provides generation code to all synthetic perturbations using parameter configs.
\end{itemize}

\subsection{Computational Requirements}
\begin{itemize}
    \item \textbf{Minimum GPU:} NVIDIA RTX 2080Ti (11GB VRAM) for inference
    \item \textbf{Recommended:} 4× A100 (80GB) for training full pipeline
    \item \textbf{Training time:} 48 hours for complete model suite on recommended hardware
    \item \textbf{Storage:} 500GB for full dataset, 100GB for core evaluation subset
\end{itemize}

\subsection{Experimental Configuration}
\begin{itemize}
    \item \textbf{Random seeds:} NumPy=42, PyTorch=42, CUDA deterministic enabled
    \item \textbf{Software versions:} PyTorch 2.0.1, CUDA 11.8, Python 3.9.16
    \item \textbf{Hyperparameter configs:} YAML files provided for all experiments
    \item \textbf{Evaluation scripts:}  Automated pipeline to give all reported metrics.
\end{itemize}

\section{Conclusion}
\label{sec:conclusion}

This study proposes a sequence-preserving, dual-FoV robust framework for traffic light and sign recognition by considering naturally prevailing perturbations in the environment. Instead of concentrating solely on adversarial degradation, the framework considers real degradations such as adverse weather (rain), dirt, physical occlusion, and glare, which are found on the road in real applications performed by autonomous vehicles.

Our main aspects are as follows: (1) a multi-source unified dataset with explicit degradation annotations of 500 sequences across four ODDs, (2) a numerical taxonomic characterization of natural perturbations, (3) a unified three-level defense stack of 79.8\% mAP attack success rate to 18.2\% attack success rate, and (4) risk-weighted metrics in relation to traffic safety.

A large number of experiments highlighted that the largest protective contribution (+2.9\% mAP) was yielded by temporal voting, and the biased perturbation due to lighting could be effectively prevented by cross-FoV validation. However, the performance for compound perturbations (rain+fog+night) was still unsatisfactory, with an accuracy of less than 50\% mAP. The inference time of the framework was 16.6ms, which allowed it to be deployed in real time at 60 frames per second (FPS) on modern GPUs (although edge deployment requires extra optimization).

However, gaps remain: limited geographic coverage (traffic data collected from US traffic systems only), vulnerability to criticized attacks (26.4\% below the threshold of 15 inflation), and poor performance with oscillation beyond 55\% occlusion. These limits indicate that, despite an increase in natural robustness, no complete safety guarantees are available.

This study provides the first insight into deployable robust perception, indicating that joint protection built using both temporal and camera redundancies offers substantial safety benefits. "However, the next generation needs to focus on system-wide gain and put new development resources towards expanding datasets globally, providing proven guarantees of robustness, and integrating with downstream planning systems if we want exposure of truly dependable autonomous driving."

\bibliographystyle{IEEEtran}   
\nocite{*}                    
\bibliography{ref}

\end{document}